\documentclass[letterpaper, 10 pt, conference]{ieeeconf}  

\IEEEoverridecommandlockouts                              %
\usepackage{caption}
\newcommand{\citep}[1]{\cite{#1}}
\usepackage{authblk}
\usepackage{subcaption}
\usepackage{wrapfig}
\usepackage{amsmath}
\usepackage{graphicx}
\usepackage{siunitx}
\usepackage{gensymb}

\usepackage{hyperref}

\usepackage{caption}
\usepackage{authblk}

\usepackage{todonotes}
\usepackage{xcolor}

\newcommand{\sect}[1]{Sec.~\ref{#1}}
\newcommand{\fig}[1]{Fig.~\ref{#1}}
\newcommand{\eq}[1]{Eq.~\eqref{#1}}

\newcommand{\tab}[1]{Table~\ref{#1}}

\usepackage{float}

\usepackage{tikz}

\newcommand{\cblock}[3]{
  \hspace{-1.5mm}
  \begin{tikzpicture}
    [
    node/.style={square, minimum size=10mm, thick, line width=0pt},
    ]
    \node[fill={rgb,255:red,#1;green,#2;blue,#3}] () [] {};
  \end{tikzpicture}%
}

\makeatletter
\newcommand{\bigplus}{%
  \DOTSB\mathop{\mathpalette\mattos@bigplus\relax}\slimits@
}
\newcommand\mattos@bigplus[2]{%
  \vcenter{\hbox{%
    \sbox\z@{$#1\sum$}%
    \resizebox{!}{0.9\dimexpr\ht\z@+\dp\z@}{\raisebox{\depth}{$\m@th#1+$}}%
  }}%
  \vphantom{\sum}%
}
\makeatother
\usepackage{amssymb}

\renewcommand{\cos}[0]{\text{c}}
\renewcommand{\sin}[0]{\text{s}}
\renewcommand{\tan}[0]{\text{t}}

\usepackage{mathtools}

\usepackage[ruled,vlined]{algorithm2e}
\usepackage{algorithmic}

\usepackage{dsfont}
\usepackage{amsmath}

\renewcommand{\vec}[1]{\boldsymbol{#1}}				

\DeclareMathOperator*{\argmax}{arg\,max}

\newcommand{\website}[0]{\url{https://sites.google.com/berkeley.edu/mbrl-ionocraft/}}

\usepackage{todonotes}
\usepackage{xcolor}

\begin{document}

%
\title{ \bf \LARGE{ Nonholonomic Yaw Control of an Underactuated Flying Robot \\ with Model-based Reinforcement Learning }
}


\author{Nathan O. Lambert\textsuperscript{1}, Craig Schindler\textsuperscript{1}, Daniel S. Drew\textsuperscript{2}, and Kristofer S. J. Pister\textsuperscript{1}
\thanks{Corresponding author: Nathan O. Lambert, \tt \href{mailto:nol@berkeley.edu}{nol@berkeley.edu}}%
\thanks{\textsuperscript{1}Department of Electrical Engineering and Computer Sciences, University of California, Berkeley.}
\thanks{\textsuperscript{2}Department of Mechanical Engineering, Stanford University.}
}



\maketitle

\begin{abstract}
Nonholonomic control is a candidate to control nonlinear systems with path-dependant states. 
We investigate an underactuated flying micro-aerial-vehicle, the ionocraft, that requires nonholonomic control in the yaw-direction for complete attitude control.
Deploying an analytical control law involves substantial engineering design and is sensitive to inaccuracy in the system model.
With specific assumptions on assembly and system dynamics, we derive a Lie bracket for yaw control of the ionocraft.
As a comparison to the significant engineering effort required for an analytic control law, we implement a data-driven model-based reinforcement learning yaw controller in a simulated flight task. 
We demonstrate that a simple model-based reinforcement learning framework can match the derived Lie bracket control -- in yaw rate and chosen actions -- in a few minutes of flight data, without a pre-defined dynamics function. 
This paper shows that learning-based approaches are useful as a tool for synthesis of nonlinear control laws previously only addressable through expert-based design.

\end{abstract}


\IEEEpeerreviewmaketitle

\section{Introduction}
\label{sec:intro}
Generating control laws for novel robots is subject to high cost-per-test and imperfect dynamics models, and these deleterious effects are amplified when deriving complex, nonlinear controllers.
For example, nonholonomic path planning has been applied to multiple underactuated systems, but requires precise knowledge of the controlled dynamical system and assumptions on resulting motion. 
In the context of novel robotics, which are often hand assembled, a perfect understanding of the dynamics is improbable.
We investigate if the performance of a novel nonholonomic control law can be matched with data-driven, reinforcement learning techniques that do not require any assumptions on dynamics.  
Model-based reinforcement learning (MBRL) has been successfully applied to a variety of robotics tasks~\cite{deisenroth2011pilco,chua2018deep, williams2017information, pmlr-v100-nagabandi20a}.
MBRL techniques iteratively plan a sequence of actions on an data-optimized dynamics model. 
In this paper we ask a question: can data-driven model-based planning effectively plan multiple steps of nonholonomic control?

Nonholonomic control is most known for the example of the parallel parking problem in automobiles -- defined by constraints limiting the path a system can take~\cite{kolmanovsky1995developments, slotine1991applied}.
Multiple applications of nonholonomic control in robotics demonstrate the potential to control these systems with closed loop, nonlinear control laws \cite{kelly1995geometric}, such as the Lie bracket sequence \cite{park1995lie}. 
Lie bracket control executes a specific, repeated control law in two planes of motion to achieve motion in an only-indirectly addressable control manifold.

\begin{figure}[t]
    \centering
    \begin{subfigure}[t]{\linewidth}
    \centering
    \includegraphics[width=0.9\linewidth]{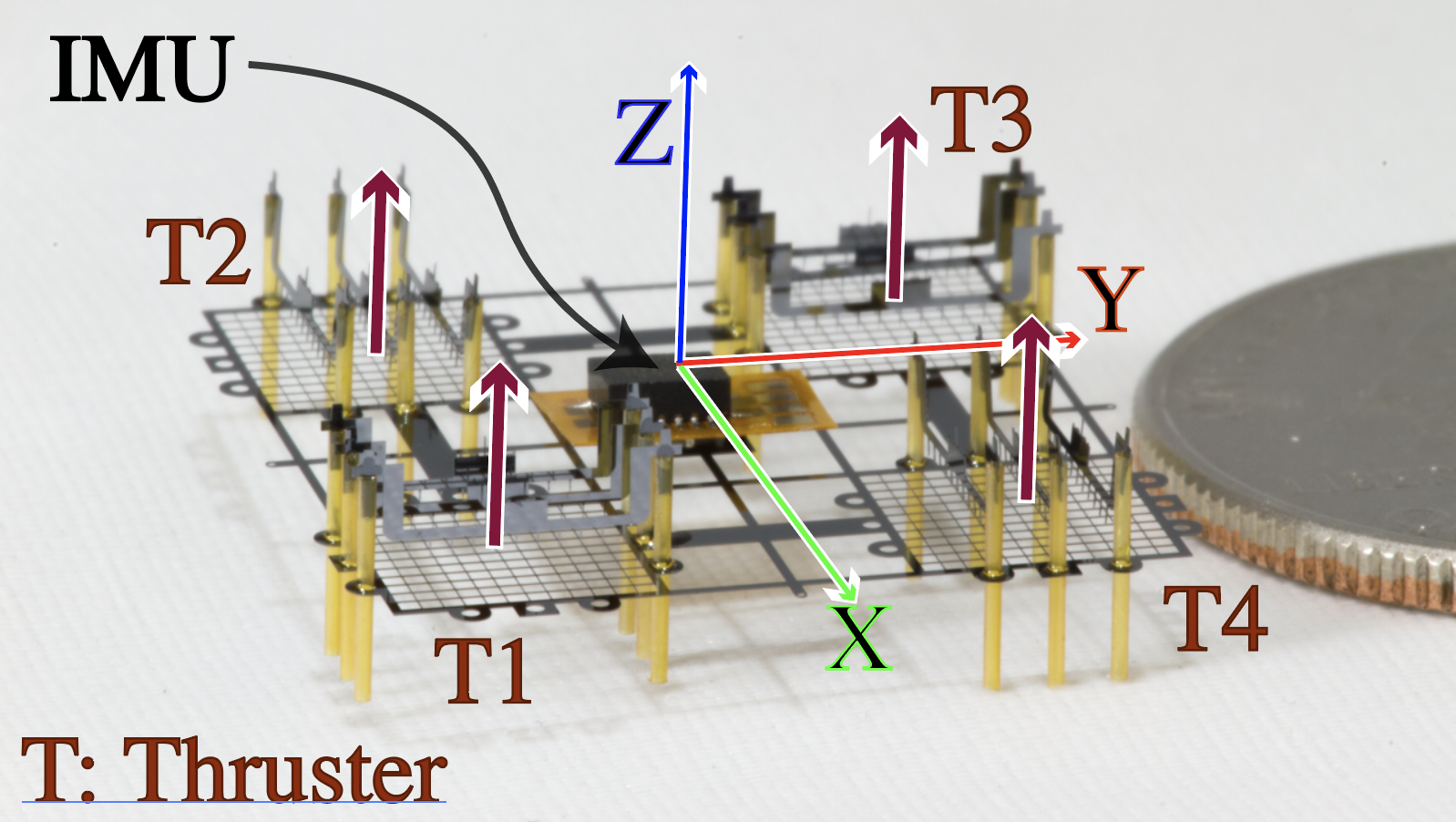}
    \caption{ 
    The studied robot is shown next to a U.S. quarter with the four thruster forces and the IMU orientation labelled.
    It masses at only \SI{30}{\milli \gram} and generates thrust from a high voltage, ``ion wind" \cite{drew2018toward}.
    }
    \label{fig:iono}
    \end{subfigure}
    \begin{subfigure}[t]{\linewidth}
    \centering
    \includegraphics[width=0.9\columnwidth]{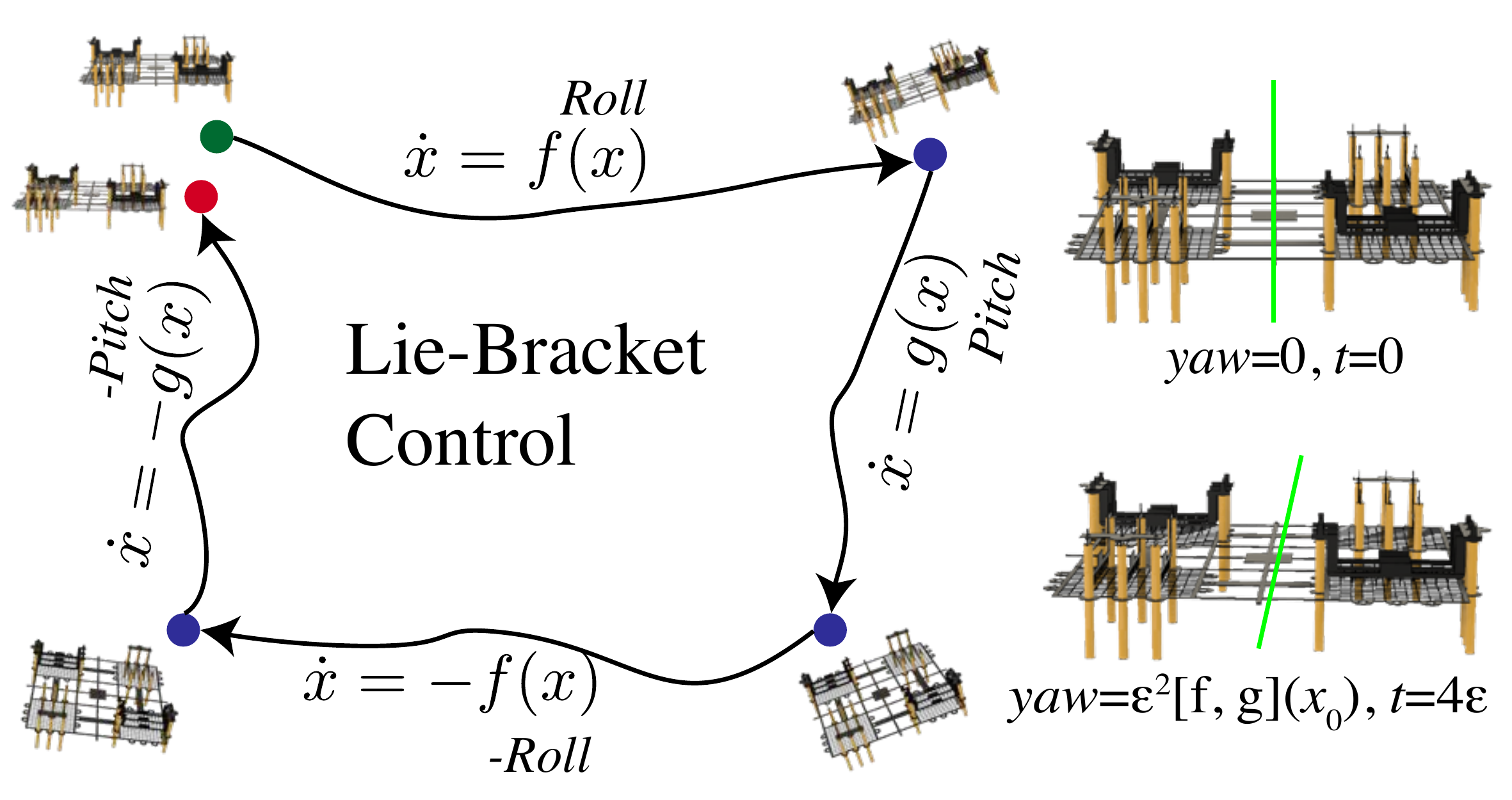}
    \caption{An illustration of Lie bracket control to produce a change in yaw.
    The Lie bracket is the manifold of motion between the starting and final position above, yielding $\text{yaw}=\epsilon^2[f,g](x_0)$ at $t=4\epsilon$.
    Our simulations show that strict Lie bracket control can achieve angular rates of up to \SI{13.2}{degrees / \second}.
    This performance is exceeded with a model-based reinforcement learning method.
    }
    \label{fig:yaw-lie}
    \end{subfigure}
    \caption{\textit{Above}: The novel robot we analyze, the ionocraft~\cite{drew2018toward}.
    \textit{Below}: The proposed nonholonomic yaw controller.
    }
    \label{fig:intro}
\end{figure}

In this paper, we contextualize our exploration of nonholonomic control and model-based planning with the ionocraft platform, a centimeter-scale flying robot with dynamics similar to a quadrotor~\cite{drew2018toward, dedic2019laser} -- \textit{i.e.} there are four individually actuatable thrusters arranged in a quadrant.
An ionocraft is an aerial vehicle which generates thrust using ionic thrusters, characterized in \cite{masuyama2013performance}. 
Quadrotors have control authority over the yaw axis via differential rotor rates that create torque about the $z$-axis, but the four-thruster ionocraft studied lacks this torque mechanism~\cite{mahony2012multirotor}.
A Lie bracket manifold exists in the yaw direction for the studied ionocraft via repeated pitch and roll motions.
The derived yaw mechanism for control via a Lie bracket approach is summarized in \fig{fig:yaw-lie}.

The contribution in this paper is to show the potential of data-driven, model-based RL to improve on 
analytical nonholonomic control in the context of a passively unstable, rapid flying robot.
Specifically, in this manuscript, we 
1) derive a new nonholonomic, Lie bracket control law for the ionocraft's yaw axis and 
2) demonstrate that model-based reinforcement learning improves on the analytical control policy in terms of yaw rate and crash frequency. 

\section{Related Works}
\label{sec:related}

\subsection{Novel Aerial Vehicles}
Development of novel actuators and small-scale electronics has progressed to multiple novel micro-aerial vehicles (MAVs)~\cite{cai2014survey, drew2017future}.
Variants on flying robots involve robots with spinning propellors, such as a tri-rotor flying robot~\cite{zou2012modeling} and others with one spinning motor~\cite{piccoli2017piccolissimo, zhang2019design}.
Other robots have shown the potential for taking advantage of the flapping wing mechanism found in nature~\cite{wood2013flight, de2016delfly}.
Robots using ion-thrusters have only recently taken off, and are yet to be controlled in experiment~\cite{drew2018toward,dedic2019laser}. 
In contrast with automobiles or walking robots, flying robots have simpler (absence of contact forces), high-speed dynamics, but are not passively stable in the sense that they are prone to crashing, resulting in a challenging control problem.

\subsection{Nonholonomic Robotic Control}
Nonholonomic control is a sub-area of nonlinear control theory studying systems with path-dependant states~\cite{kolmanovsky1995developments, slotine1991applied}. 
One method of nonholonomic control uses the Lie bracket, which involves repeated, mirrored actuation on two input manifolds~\cite{park1995lie}, as detailed in \sect{sec:lie_formula}.
Nonholonomic control is frequently applied to wheeled robots where the plane of motion must follow the direction of the wheels, leading to the parallel parking problem \cite{paromtchik1996autonomous}.
An extension of the 4-wheeled parallel parking problem is the controllability of two-wheeled balancing robots~\cite{salerno2003nonlinear}.
A novel trident snake robot uses nonholonomic control to maneuver with three arms attached to non-slide wheels~\cite{ishikawa2005trident}.

There are limited studies of nonholonomic control applied to flying robots. 
Nonholonomic path planning is used for simulated flying vehicles~\cite{masehian2015path}, but it does not consider details of low-level dynamics needed for attitude control.
A space-robot with a multi-joint arm attached to a main body uses nonholonomic control with a Lyapunav function for path-planning \cite{nakamura1990nonholonomic} and nonholonomic redundancy to avoid joint limits \cite{nakamura1993exploiting}.
Another contactless-medium of motion is swimming, and nonholonomic control is used to imitate carangiform swimming of a planar, rigid fish-robot \cite{morgansen2001nonlinear} and to control a AUV~\cite{zain2011nonholonomic}. 
A novel nonholonomic controller called ``wriggle-steering'' uses repeated and mirrored sinusoidal oscillations -- like a Lie bracket in a frequency domain --  to yaw a micro-flapping wing robot \cite{fuller2015rotating}.
Our work extends the application of nonholonomic, Lie bracket control to that of low-level control of a flying rigid-body, with challenging, rapid dynamics.

\subsection{Model-based Reinforcement Learning}
Model-based reinforcement learning (MBRL) has emerged as a functional candidate for robotic control in a data-efficient manner~\cite{deisenroth2011pilco,williams2017information, chua2018deep, janner2019trust} -- this paper extends the functionality of MBRL to nonholonomic planning for flying robots.
Visual MBRL (\textit{i.e.} control from pixels) has been shown to work in a simple, simulated nonholomic car task with Linear Quadratic Regulator (LQR) control and variational inference \cite{zhang2019solar}.
Additionally, MBRL has been shown to learn to control a micro-quadrotor with no assumption on dynamics~\cite{lambert2019low} and with suspended payloads modifying dynamics~\cite{belkhale2020model}.
MBRL for attitude control of quadrotors is encouraging for data-driven control of a novel MAV, and this work extends the results to show the capability of MBRL algorithms to plan complex multi-step action sequences for flying robots.

\section{Background}

\subsection{Lie Bracket Control}
\label{sec:lie_formula}
For a nonlinear, dynamical system with two inputs, $\vec{v} = \begin{bmatrix} v_1 & v_2 \end{bmatrix}$, a Lie bracket control law is defined by a repeated sequence of motions for a small time, $\epsilon$, to generate motion in a new manifold of motion, $\big[ f,g \big]$, called the Lie bracket vector field. 
Consider the following system,
\begin{equation}
    \dot{x} = f(x)v_1 + g(x) v_2
    \label{eq:form}
\end{equation}
The Lie bracket control law is motivated by a series of Taylor series expansions along the vector fields of each input $f, g$, so that a third manifold of motion emerges, $\big[ f,g \big]$.
Specifically, consider the input sequence below,
\begin{equation}
    v(t)  =
 \begin{cases} 
    [1, 0] & t\in[0,\epsilon) \\
    [0, 1] & t\in[\epsilon,2\epsilon)\\
    [-1, 0] & t\in[2\epsilon,3\epsilon) \\
    [0, -1]  & t\in[3\epsilon,4\epsilon)
    \end{cases}
    \label{eq:lie-control}
\end{equation}
After $t=4\epsilon$, the state can be expressed as (shown in~\cite{khalil2002nonlinear})
\begin{align}
    x(4\epsilon) &= x_0 + \epsilon^2 [f,g](x_0)+\mathcal{O}(\epsilon^3), \\
    \text{where} & \quad \big[ f,  g \big] = \frac{df}{dx}g - \frac{dg}{dx}f.
    \label{eq:lie}
\end{align}
For a visualization of the Lie bracket motion in a phase plane caused by two inputs, see \fig{fig:yaw-lie}.

\subsection{Model-based Reinforcement Learning}
Model-based reinforcement learning (MBRL) follows the framework of an agent interacting in an environment, learning a model of said environment, and then leveraging the model for control~\cite{sutton2018reinforcement}.
Specifically, the agent acts in a Markov Decision Process (MDP) governed by a transition function $x_{t+1} = f(x_t,u_t)$ and returns a reward at each step $r(x_t,u_t)$.
With a collected dataset $\mathcal{D} \coloneqq \{x_i, u_i, x_{i+1},r_i\}_{i=1}^N$, the agent learns a model, $x_{t+1} = f_\theta(x_t,u_t)$ to minimize the negative log-likelihood of the transitions. 
We employ sample-based model-predictive control (MPC) using the learned dynamics model, which optimizes the expected reward over a finite, recursively predicted horizon, $\tau$, from a set of actions sampled from a uniform distribution $\mathcal{U}(a)$, as
\begin{align}
    u^* & = \argmax_{u_{t:t+\tau}} \sum_{i=0}^{\tau} r(\hat{x}_{t+i}, u_{t+i}), \\
    s.t. & \quad   \hat{x}_{t+1}   = f_\theta(\hat{x}_t,u_t).
    \label{eq:mpc}
\end{align}

\section{Ionocraft Model}
\label{sec:iono}

In this paper, we model and study an ionocraft~\cite{drew2018toward, dedic2019laser}; a flying robot with four individually-addressable electrohydrodynamic thrusters, as shown in \fig{fig:iono}.
The robot is approximately \SI{4}{cm^2} and masses \SI{30}{\milli \gram} not including the sensor payload.
Electrohydrodynamic thrust, summarized in \cite{masuyama2013performance}, is produced by the momentum-transferring collisions between ions drifting in an applied electric field and the particles in a neutral fluid.
The forces, $F_k(i_k,\beta_k)$, are models for the electrohydrodynamic force from each thruster~\cite{drew2018toward}, as
\begin{equation}
    F_{ion,k} = \beta_k \frac{i_{k}d}{\mu}.
    \label{eq:thrust}
\end{equation}
Here, $i$ is the ion current, $d=500\mu m$ is the air-gap and $\mu=2\cdot 10^{-4} m^2/V$ is the ion mobility of $N_2^+$ in dry air. 
Fitting a $\beta_k$, which represents a deviation in force from that expected by an idealized one-dimensional derivation, to each thruster allows more flexibility in model fitting and simulation. 
Past work shows $\beta \in [0.5, 0.7]$ for the region of currents $i \in [0,.5]mA$ corresponding to voltages $v \in [1650,2100] V$~\cite{drew2018toward}.

The ionocraft dynamics model follows a $12$ state rigid-body dynamics with a state $x=[X \ Y \ Z \ \psi \ \theta \ \phi \ v_x \ v_y \ v_z \ \omega_x \ \omega_y \ \omega_z]$, and input $u=[F_1 \ F_2 \ F_3 \ F_4]$\SI{}{\milli \newton}, centered around an unstable equilibrium point $x^*=\vec{0}$, which are simulated at a dynamics frequency of \SI{1}{k \hertz}, a control frequency of \SI{100}{\hertz} to match the experimental setup used in~\cite{drew2018ionocraft}, and Gaussian state noise in each dimension, $\sigma \sim \mathcal{N}(0,0.01)$:
\begin{align}
    \begin{bmatrix}
    \dot{X} \\
    \dot{Y} \\
    \dot{Z} 
    \end{bmatrix} & =
    Q_{B/I}
    \begin{bmatrix}
    v_x \\
    v_y \\
    v_z 
    \end{bmatrix}, \\
    \begin{bmatrix}
    \dot{\psi} \\
    \dot{\theta} \\
    \dot{\phi} 
    \end{bmatrix} & =
    W^{-1}
    \begin{bmatrix}
    \omega_x \\
    \omega_y \\
    \omega_z 
    \end{bmatrix}, \\
    \begin{bmatrix}
    \dot{v}_x \\
    \dot{v}_y \\
    \dot{v}_z
    \end{bmatrix} & =
    \frac{1}{m}
    \begin{bmatrix}
    F_x \\
    F_y \\
    F_z 
    \end{bmatrix} 
    -
    \begin{bmatrix}
    0 & -\omega_z & \omega_y\\
    \omega_z & 0 & -\omega_x \\
    -\omega_y & \omega_x & 0 
    \end{bmatrix}
    \begin{bmatrix}
    v_x \\
    v_y \\
    v_z 
    \end{bmatrix}, \\
    I_B
    \begin{bmatrix}
    \dot{\omega}_x \\
    \dot{\omega}_y \\
    \dot{\omega}_z
    \end{bmatrix} & =
    \begin{bmatrix}
    \tau_x \\
    \tau_y \\
    \tau_z 
    \end{bmatrix} 
    -
    \begin{bmatrix}
    0 & -\omega_z & \omega_y\\
    \omega_z & 0 & -\omega_x \\
    -\omega_y & \omega_x & 0 
    \end{bmatrix}
    I_B
    \begin{bmatrix}
    \omega_x \\
    \omega_y \\
    \omega_z 
    \end{bmatrix} 
    \label{eq:dynam}
\end{align}
The inverse Wronskian $W^{-1}\in\mathbb{R}^{3\times3}$ and the body-to-inertial transformation $Q_{B/I}\in\mathbb{R}^{3\times3}$ defined with the trigonometric functions abbreviated as $ \cos \coloneqq \text{cos}$,  $\sin  \coloneqq \text{sin} $, and $\tan \coloneqq \text{tan} $:
\begin{align}
     Q_{B/I} & = \begin{bmatrix}
    \cos(\theta )\cos(\psi ) & \cos(\psi )\sin(\theta )\sin(\phi )  & \sin(\phi )\sin(\psi ) \\
    & - \cos(\phi )\sin(\psi ) & + \cos(\phi )\cos(\psi )\sin(\theta ) \\
    \cos(\theta )\sin(\psi ) & \cos(\phi )\cos(\psi )& \cos(\phi )\sin(\theta )\sin(\psi )  \\
    & + \sin(\theta )\sin(\phi )\sin(\psi ) & - \cos(\psi )\sin(\phi ) \\
     -\sin(\theta )& \cos(\theta )\sin(\phi ) & \cos(\theta )\cos(\phi ) 
    \end{bmatrix} \\
    W^{-1} & = \frac{1}{\cos(\theta)} \begin{bmatrix}
    0 & \sin(\phi) & \cos(\phi) \\
    0 & \cos(\phi) \cos(\theta) & - \sin(\phi) \cos(\theta) \\
    \cos(\theta) & \sin(\phi) \sin(\theta) & \cos(\phi) \sin(\theta)
    \end{bmatrix}.
\end{align}
The simulator emulates low-level control by actuating the ionocraft with voltages, which map to currents and forces via \eq{eq:thrust}. 
Then, the rigid-body model incorporates torques and forces as an ideal point-force from the center of the thruster, $l=$\SI{1}{\centi \meter}, from the center-of-mass.
\begin{align} 
    \left[ 
	\begin{array}{c}
    	F_z \\
        \tau_z \\
        \tau_y \\
        \tau_x
    \end{array}
	\right]
    = \left[
    \begin{array}{cccc}
    	1 & 1 & 1 & 1\\
        0 & 0 & 0 & 0 \\
        -l & -l & l & l \\
        -l & l & l & -l
    \end{array}
    \right]
    \left[
    \begin{array}{c}
    F(i_1, \beta) \\
     F(i_2, \beta) \\
     F(i_3, \beta) \\
     F(i_4, \beta)
    \end{array}
    \right] .
\label{eq:forces}
\end{align}

The important observation for yaw control is that the $z$-axis torque is totally decoupled from the individual thrusters.
Neglecting the damping forces in low velocity maneuvers such as hovering, a corresponding subset of the Jacobian linearization of the ionocraft approximated around equilibrium, $x^* = \vec{0}$, is 
\begin{align}
    \dot{\psi} &= \omega_z, & \dot{\theta} &= \omega_y,  & \dot{\phi} &= \omega_x+ \omega_z, \\
    \dot{\omega_x} &= \frac{\tau_x}{I_{xx}},  &
    \dot{\omega_y} &= \frac{\tau_y}{I_{yy}},  &
    \dot{\omega_z} &= \frac{\tau_z}{I_{zz}}.  
    \label{eq:linearized}
\end{align}
A direct way to see a loss of yaw-actuation is with a linearized controllability matrix, $\tilde{W_c}(\widetilde{A},\widetilde{B})$, 
\begin{align}
\tilde{W_c} = \left[
\begin{matrix}
\tilde{B} & \tilde{A} \tilde{B} & \dots & \tilde{A}^{n-1} \tilde{B}
\end{matrix}
\right] .
\label{eq:matcontrol}
\end{align}
A robot can actuate all directions if this matrix is full rank. 
The matrix $\tilde{W_c}$ is of rank $n-1$, with a corresponding uncontrollable mode of $\dot{\omega_z}$, but this also propagates to $\dot{\psi}$.
This loss of controllability formally shows the need for a nonlinear yaw control law. 

\section{Nonholonomic Yaw Control}
In this section we derive the analytical Lie bracket controller and compare it to a data-driven MBRL approach for yaw control.
With both methods, the goal of the approach is to maximize the yaw, $\psi$, without causing substantial flight disturbance (\textit{e.g.} reaching the stop condition in the simulator of the magnitude of pitch, $\theta$, or roll, $\phi$, above \SI{45}{degrees}).
Code, video, and an expanded manuscript are available on the website: \website{}.

\subsection{Analytical Control: Lie Bracket}
\label{sec:lie}

Here we map the Lie bracket formulation outlined in \sect{sec:lie_formula} to an open-loop yaw controller for the ionocraft. 
First, consider the Euler rates defined in \eq{eq:dynam}, with the trigonometric functions abbreviated.
\begin{align}
    \dot{\psi} &= \frac{\sin(\phi)}{\cos(\theta)}\omega_y + \frac{\cos(\phi)}{\cos(\theta)}\omega_z, \label{eq:euler_rate3}\\
    \dot{\theta} &= \cos(\phi)\omega_y - \sin(\phi) \omega_z, \label{eq:euler_rate2} \\
    \dot{\phi} &= \omega_x+\sin(\phi)\tan(\theta)\omega_y + \cos(\phi)\omega_z.
    \label{eq:euler_rate}
\end{align}
To study as nonlinear control, expand the differential equations for Euler rates with angular rates as inputs. 
In order to assume the angular rates are inputs, we must show that they are directly and independently actuated.
To do so, consider the angular rate differential equations shown below, assuming the inertial matrix is diagonal: $I_B = \operatorname{diag}(I_{xx},I_{yy},I_{zz})$. 
\begin{align}
    \dot{\omega_x} &= \frac{1}{I_{xx}}\big( \tau_x + \omega_z \omega_y I_{yy} - \omega_y \omega_z I_{zz}\big), \\
    \dot{\omega_y} &= \frac{1}{I_{yy}}\big(\tau_y + \omega_z \omega_x I_{xx} - \omega_x \omega_z I_{zz}\big), \\
    \dot{\omega_z} &= \frac{1}{I_{zz}}\big(\tau_z + \omega_y \omega_x I_{xx} - \omega_x \omega_y I_{yy}  \big).
    \label{eq:angular_rate}
\end{align}
The following assumptions are needed to prove the existence of the Lie bracket yaw controller:
\begin{enumerate}
    \item For a symmetric body in the $XY$-plane,  $I_{xx}=I_{yy}$, so the cross terms of $\dot{\omega_z}$ are equal, $\omega_y \omega_x I_{xx} = \omega_x \omega_y I_{yy}$.
    \item  The robot starts from rest, $\omega_z(t_0)=0$, so the cross terms in $\dot{\omega_x}, \dot{\omega_y}$ are zero initially.
    \item With the lack of yaw coupling in thrusters, the robot applies $0$ torque about the $z$ axis, $\tau_z=0$. 
\end{enumerate}
Combining assertions 2) and 3) enforces $\omega_z=0 \forall t$. 
With 1), this means that $\dot{\omega_x}, \dot{\omega_y}$ are fully determined by $\tau_x, \tau_y$, so we can model the flight dynamics as a nonlinear system with two manifolds of action, as in \eq{eq:form}.

Therefore, for the flying robot define the state space with the following state variables and vector fields:
\begin{equation}
    \vec{x} = \begin{bmatrix}
    \psi & \theta & \phi 
    \end{bmatrix},
    \quad ~
    \vec{v} = \begin{bmatrix}
    \omega_{x} & \omega_{y} & \omega_{z}
    \end{bmatrix},
\end{equation}
\begin{equation}
    \begin{bmatrix}
    \dot{\psi} \\
    \dot{\theta} \\
    \dot{\phi} 
    \end{bmatrix} = f(x) \omega_x +g(x) \omega_y +h(x) \omega_z .
\end{equation}
Define the vector fields from  \eq{eq:euler_rate3}, \eq{eq:euler_rate2}, and \eq{eq:euler_rate}.
\begin{equation}
    f(x) = \begin{bmatrix}
    0 \\
    0 \\
    1 \\
    \end{bmatrix}
    g(x) = \begin{bmatrix}
    \frac{\sin (\phi)}{\cos (\theta)} \\
    \cos (\phi) \\
    \sin(\phi)\tan(\theta) \\
    \end{bmatrix}
    h(x) = \begin{bmatrix}
    \frac{\cos (\phi)}{\cos (\theta)}\\
    -\sin (\phi) \\
    \cos(\phi) \\
    \end{bmatrix}.
\end{equation}
Making the assumptions above guarantees that $h(x)$ does not affect the differential equation, and there is no direct yaw actuation. 
Consider two Lie brackets $\big[ f, g \big]$ and $\big[ g, f \big]$, as defined in \eq{eq:lie}. 

Computing the Jacobian of our dynamics $f$ and $g$ yields,
\begin{equation}
    \frac{df}{dx} =
    0_{3 \times 3}, \quad
    \frac{dg}{dx} =
    \begin{bmatrix}
    0 & \frac{2\sin(\phi)\sin(\theta)}{\cos(2\theta)+1} & \frac{\cos (\phi)}{\cos (\theta)}\\
    0 & 0 & -\sin(\phi)\\
    0 & \sin(\phi)\cos^2(\theta) & \cos (\phi)\tan (\theta)\\
    \end{bmatrix}.
\end{equation}
Finally, the Lie bracket is computed as: 
\begin{equation}
    \big[ f,  g \big](x) = - \frac{dg}{dx}f(x) = -
    \begin{bmatrix}
     -\frac{\cos (\phi)}{\cos (\theta)}\\
     \sin(\phi)\\
     - \cos (\phi)\tan (\theta)\\
    \end{bmatrix} .
\end{equation}
The final assumption is that the control law will be applied near hover, $\theta, \phi \approx 0$.
This approximation shows controllability in the yaw-direction with low pitch or roll drift captured in the higher order terms, $\mathcal{O}(\epsilon^3)$ in \eq{eq:lie}  -- completing attitude control for the ionocraft. 
\begin{equation}
    \big[ f,  g \big](x) \Big|_{\theta=0,\phi=0} 
    =
    \begin{bmatrix}
     1\\
     0\\
     0\\
    \end{bmatrix} 
    \label{eq:yaw_motion}
\end{equation}
Intuitively, the robot actuates yaw by repeated, mirrored pitch-roll oscillations, but the assumptions on symmetry are highly sensitive to the reality of imperfect assembly outlined in \tab{tab:params2}.

\begin{table}[t]
\vspace{4pt}
\begin{minipage}{.32\linewidth}
\begin{center}
 \begin{tabular}{ l c} 
   \multicolumn{2}{c}{No IMU}  \\ [0.5ex] 
 \hline 
  Mass & 26mg  \\
  $I_{XX}$ & $1.967$\\ 
  $I_{YY}$ & $1.967$\\
  $I_{ZZ}$ & $3.775$\\
  [1.0ex] 
\end{tabular}
\end{center}
\end{minipage}
\begin{minipage}{.32\linewidth}
\begin{center}
 \begin{tabular}{ l c} 
   \multicolumn{2}{c}{IMU Center}  \\ [0.5ex] 
 \hline 
  Mass & 46mg  \\
  $I_{XX}$ & $1.984$ \\
  $I_{YY}$ & $1.983$ \\
  $I_{ZZ}$ & $3.804$ \\
   [1.0ex] 
  
\end{tabular}

\end{center}
\end{minipage}
\begin{minipage}{.32\linewidth}
\begin{center}
 \begin{tabular}{ l c} 
   \multicolumn{2}{c}{IMU 5mm x-error}  \\ [0.5ex] 
 \hline 
  Mass & 46mg  \\
  $I_{XX}$ & $2.262$ \\
  $I_{YY}$ & $1.983$ \\
  $I_{ZZ}$ & $4.083$ \\ [1.0ex] 
  
\end{tabular}

\end{center}
\end{minipage}
\caption{Different assembly configurations and parameters showing the change of inertial properties from fabrication error in IMU placement -- breaking the symmetric body assumptions needed to safely apply Lie bracket yaw control.
The inertial values are reported in \SI{}{\gram \per \milli \meter^2} and the cross terms are small, \textit{i.e.} $\{I_{XY},I_{XZ},\cdots\}<.01$.
}
\label{tab:params2}
\end{table}

Translating the input vector, $v$, used in the Lie bracket derivation to an input of motor forces, $u$, is done by setting and holding specific angular accelerations.
Given a desired angular rate, the controller can set the angular rate with a simple force-to-torque model.
For example, the angular acceleration about the $x$-axis follows: $\dot{\omega}_x = \frac{l}{I_{XX}}\cdot(F_1+F_2-F_3-F_4)$, where $l=$\SI{1}{\centi \meter} is the length from the center of mass to the center of the thruster.  
The control setup and simulator use specific application of motor forces rather than setting an desired angular rate, so 
we define the actions used by the Lie bracket sequence to yaw the robot in \eq{eq:lie-actions}. 
Applying a inner-loop controller to set the angular rates would aid the Lie bracket controller and aid with fabrication variance, but the current design of the robot does not have this capability.
Consider $\vec{u}^x_y$, where $x$ corresponds to a direction $\in\{+,-\}$ of the dimension $y\in\{\text{pitch},\text{roll}, \text{equil.}\}$. 
\begin{align}
    &  \vec{u}^+_\text{pitch}= [0.15,0.05,0.05,0.15]\ \SI{}{\milli \newton} \nonumber \\
    &\vec{u}^+_\text{roll}= [0.15,0.15,0.05,0.05]\ \SI{}{\milli \newton} \nonumber\\
    &\vec{u}^-_\text{pitch} = [0.05,0.15,0.15,0.05]\ \SI{}{\milli \newton} \nonumber \\
    &\vec{u}^-_\text{roll}= [0.05,0.05,0.15,0.15]\ \SI{}{\milli \newton}  \nonumber\\
     &  \vec{u}_\text{equil}=[0.1,0.1,0.1,0.1]\ \SI{}{\milli \newton}
    \label{eq:lie-actions}
\end{align} 
These actions are designed given flight mass of 50mg to assume a payload such as an IMU and a thrust coefficient, $\beta=0.6$ (equilibrium is the action with all motors set to output of $F_i = \frac{mg}{4}$).
A hard-coded sequence of these actions at \SI{100}{\hertz} ($\epsilon=0.01$) achieves a baseline yaw rate of \SI{1}{degrees / \second}. 
This yaw rate can be increased by a corresponding increase in the time that each action is held, $\epsilon$.
\tab{tab:lie-rates} shows that the raw rate can increase with the Lie bracket controller up to \SI{13.2}{degrees / \second}, but faster yaw actuation causes faster instability in pitch and roll.
To slow divergence and get peak yaw-rate with the Lie bracket, we remove all state noise, $\sigma=0$, and zero the initial state, $\vec{x}(t_0)= 0$. 


Translating control methods to robotic hardware strongly correlates with the practical value of a method for control synthesis.
Lie bracket control is based on careful assumptions of symmetry in movement and near-equilibrium flight, which can easily break down due to environmental disturbances or nonideality in robot assembly.
Specifically, the ionocraft needs an on-board MPU-9250 inertial measurement device (IMU) placed in the center for state-space measurement. 
This assembly is done by hand and subject to errors, which can have substantial alteration on the inertial properties of the robot -- summarized in \tab{tab:params2}.
The hand-assembly also conveys an increased cost-per-test, so the ideal method for generating a controller would be able to generalize across inaccurate dynamics and be sample efficient.

\begin{table}[t]
\centering
\vspace{4pt}
 \begin{tabular}{c|cc} 
  Lie Bracket $\epsilon$ (s) & Yaw Rate (\SI{}{deg. / \second}) & Simulation Stop (s)  \\
   \hline 
0.01 & 2.0  & 7.52\\ 
0.02 & 3.9 & 3.78 \\ 
0.03 & 5.9 & 2.51  \\ 
0.04 & 7.9 & 1.88  \\ 
0.06 & 11.5 & 1.30  \\ 
0.08 & 13.2 & 0.94  \\ 
\end{tabular}
\caption{The different yaw rates achieved by a Lie bracket controller with different values of the time-hold, $\epsilon$, for each action in the sequence.
Without any feedback control on pitch or roll, the Lie bracket controller diverges and reaches the stop condition of pitch or roll greater than \SI{45}{degrees}.
There is a quadratic increase in yaw-rate corresponding to the time-length each action in the sequence (following \eq{eq:lie}) $\vec{u}^+_\text{pitch}, \ \vec{u}^+_\text{roll}, \ \vec{u}^-_\text{pitch}, \ \vec{u}^-_\text{roll}$, at a cost of faster divergence of attitude stability.
}
  \label{tab:lie-rates}

\end{table}



\subsection{Data-driven Control: MBRL}

The main shortcoming of analytical methods is the strong assumptions on system dynamics, so we explore model-based reinforcement learning.
While free from assumptions on system dynamics, the MBRL framework 
has a new cost of environmental interactions that we quantify and consider.
The predictive dynamics model is a simple feed-forward neural network with two hidden layers, trained with the Adam optimizer for 17 epochs with a learning rate of $0.0025$ and a batch size of 18. 
A parameter that can strongly affect performance is the prediction horizon of the MPC, and in this paper, we use a horizon of $\tau=5$.
The number of evaluated actions is $N=500$ and the actions for each thruster are sampled from uniform force distributions $F_i \in \mathcal{U}(0, 0.3) \SI{}{\milli \newton}$, as
\begin{equation}
    u_i = [F_1 \ F_2 \ F_3 \ F_4] \SI{}{\milli \newton}.
\end{equation}
In this section we detail the design decisions that must be made to achieve nonholonomic behavior with a model-based controller, which primarily entails generating a suitable reward function.

A simple formulation for achieving yaw in a reinforcement learning framework would be to encourage maximizing the magnitude of yaw, such as $r(\vec{x},\vec{u}) = \psi$, but this reward function does not penalize a loss of attitude stability.
In practice, this reward function quickly diverges and hits the simulator pitch, roll stop condition of $|\theta| > $ \SI{45}{degrees} or $|\phi| >$ \SI{45}{degrees}. 
To mitigate this, the objective becomes a dual optimization with the reward function balancing the reward of yaw versus the cost of pitch and roll. 
Simply appending a scaled squared cost on pitch and roll (\textit{e.g.} $r(\vec{x},\vec{u}) = \psi^2 - \lambda(\theta^2 + \phi^2)$) does not achieve stable yaw control because as yaw increases globally, the controller accepts a corresponding increase in pitch and roll to increase reward -- eventually the robot reaches high angles and ends the simulation early in $30\%$ of trials, even when trained on minutes of flight data.


\begin{figure}[t]
    \vspace{8pt}
    \begin{center}
    \small{\cblock{0}{0}{200} Roll  (\textcolor[rgb]{.01,.01,.71}{$\bigplus$})\quad
    \cblock{200}{130}{0} Pitch
    (\textcolor[rgb]{.78,.6,.01}{\large{$\circ$}}) \quad
    \cblock{10}{200}{20} Yaw
    (\textcolor[rgb]{.1,.8,.2}{$\mathbb{X}$})
    } 
    \end{center}
    \centering
     \includegraphics[width=\columnwidth]{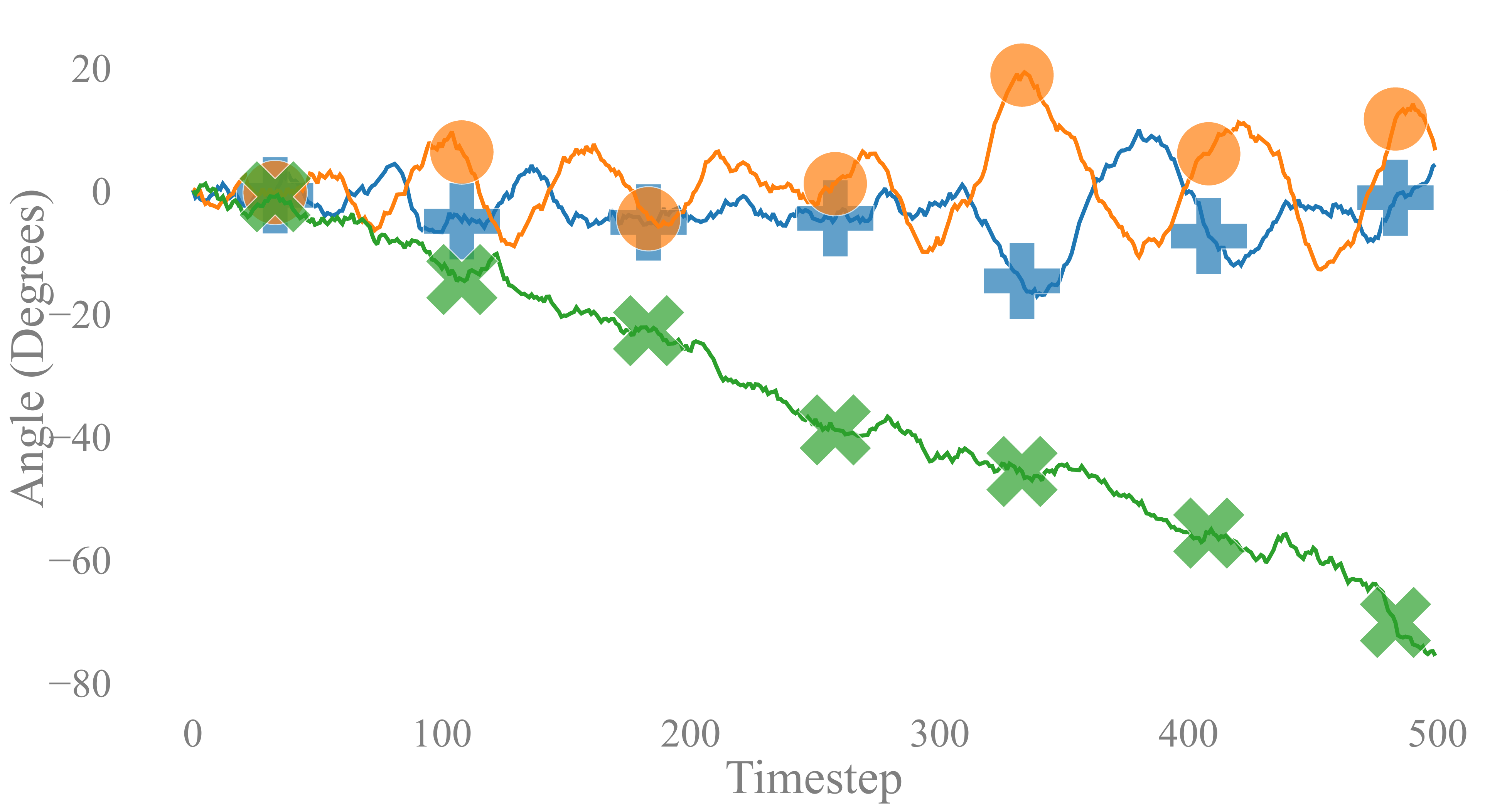}
    \caption{The Euler angle response with MBRL-MPC of the ionocraft when maximizing the yaw rate. 
    The out of phase pitch and roll oscillations mirror that of the Lie bracket control, where directly actuating pitch and roll is needed to generate movement in the Yaw plane.
    %
    }
    \label{fig:yaw-response}
\end{figure}

To alleviate this, sliding mode control~\cite{edwards1998sliding} can ensure attitude stability while maximizing yaw rate when in a stable region of the state space, as
\begin{equation}
    r(\vec{x},\vec{u}) = \left\{ \begin{array}{cc}
        |\psi| & (|\phi|<\eta) \cap (|\theta| <\eta), \\
         - (\theta^2 + \phi^2) & else.
    \end{array} \right. 
\end{equation}
This introduces a tune-able hyperparameter, the attitude window, $\eta$, weighting the balance between attitude stability and yaw rate that was set to \SI{10}{degrees} during our experiments. 
With this reward function, agents never reach the simulation stop condition when trained on at least \SI{20}{\second} of flight data.

\begin{figure}[t]
\vspace{4pt}
\begin{center}
    \small{\cblock{0}{0}{200} Environment Variation   (\textcolor[rgb]{.01,.01,.71}{$\bigplus$})\quad
    \cblock{220}{170}{0} No Variation
    (\cblock{220}{170}{0}) }
    \end{center}
    \begin{center}
    \small{
    \cblock{10}{10}{10} Ideal Lie Bracket, $\epsilon = 0.08\text{s}$, 
    (\textcolor[rgb]{.1,.1,.1}{- - -})
    }
    \end{center}
    \centering
    \includegraphics[width=\columnwidth]{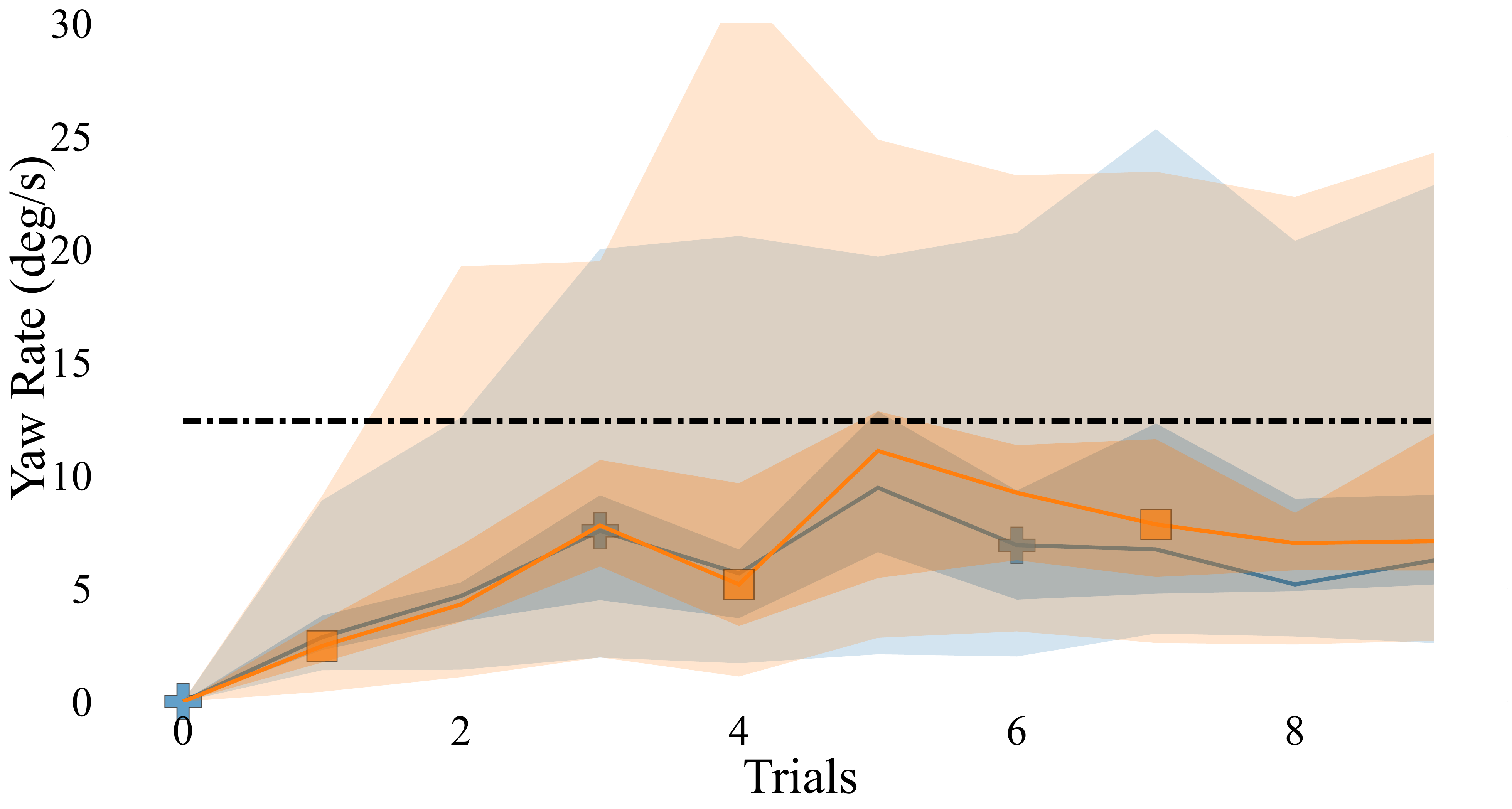}
    \caption{
    The median reward ($N_\text{robot}=25$), with $65^{th},95^{th}$ percentiles shaded, per trial when learning yaw control for the simulated ionocraft with a $15\%$ variation on inertial properties and a wide range of initial states.
    The agent with environmental variation closely matches the one with symmetric dynamics and stable initial conditions, highlighting the ability for MBRL to cope with realistic hardware variation.
    One trial is equal to 1000 environment steps.
    The agent learns to stably actuate yaw, with many trials with a yaw rate over the peak derived Lie bracket at \SI{13.2}{degrees \per \second}. 
    }
    \label{fig:yaw-learn}
\end{figure}

In order to test the robustness of the MBRL approach to manufacturing and environmental disturbances that substantially degrade the performance of the Lie bracket control law, we randomized the inertial properties to $\pm 15\%$ the nominal value and sample initial Euler angles $\phi,\theta \in [-22.5,22.5]\text{degrees}$.
The performance with this randomization matches closely to that with initial state set to zero, $\vec{x}_0 = 0$, and symmetric dynamics, showing the ability for the MBRL approach to adapt to variable environments, shown in \fig{fig:yaw-learn}.
Each trial represents a maximum of 1000 training steps, so the full training set after 10 trials represents just \SI{100}{\second} of training data.
The learning agents learn (with and without assembly and environmental variation) mean behavior just below the Lie bracket baseline (in variation free environment), and the MBRL agents have many trials with a substantially greater yaw rate.
A simulated flight, where a MBRL agent uses pitch and roll oscillations similar to a Lie bracket control law is shown in \fig{fig:yaw-response}.

\subsection{Learning to Mimic the Lie bracket}
\begin{figure}[t]
\vspace{4pt}
    \begin{center}
    \small{\cblock{0}{0}{200} MBRL Actions   (\textcolor[rgb]{.01,.01,.71}{$\bigplus$})\quad
    \cblock{200}{130}{0} Lie bracket ($\epsilon=0.05$)
    (\textcolor[rgb]{.78,.6,.01}{\large{$\circ$}}) 
    } 
    \end{center}
    \centering
     \includegraphics[width=\columnwidth]{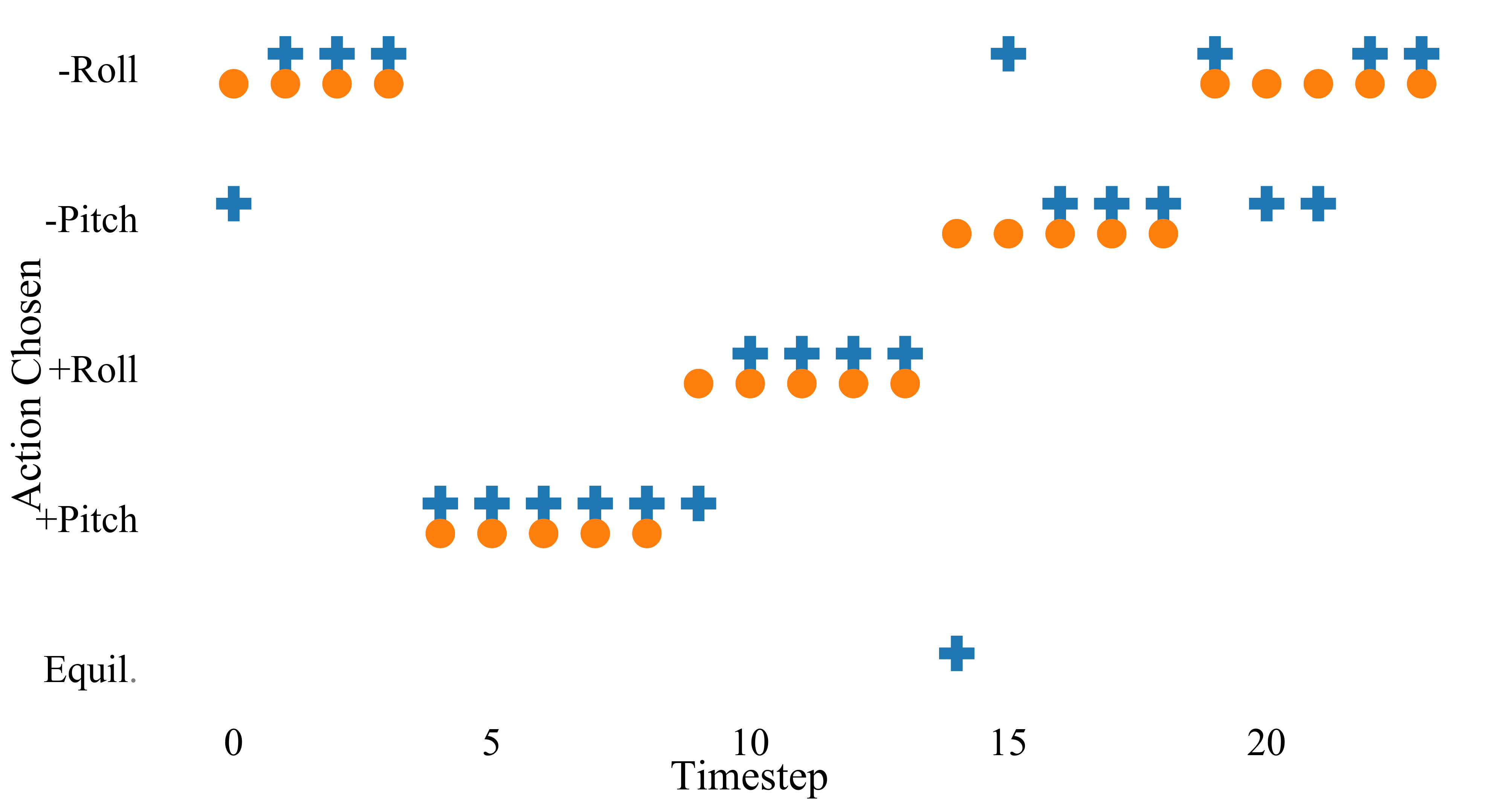}
    \caption{A comparison of the chosen actions when the learning-based approach can only choose from the actions used in the Lie bracket from \sect{sec:lie} and an equilibrium action.
    The actions represent the first 25 actions of a trajectory from a nonzero initial pitch and roll with symmetric inertial matrices using the sampling-based MPC.}
    \label{fig:yaw-actions}
\end{figure}

An important point when considering the learned behavior is if it mimics the Lie bracket control, or does the MBRL algorithm develop unintended, simulation-exploiting maneuvers.
To test this, we train \emph{a more constrained version} of the MBRL algorithm with a constrained action space of the actions used in Lie bracket control and equilibrium: 
\begin{equation}
    u\sim [\vec{u}^+_\text{pitch}, \ \vec{u}^+_\text{roll}, \ \vec{u}^-_\text{pitch}, \ \vec{u}^-_\text{roll}, \ \vec{u}_{\text{equil}}].
\end{equation}
The standard Lie bracket formulation actuates each action for one time step to mitigate divergence of other states from higher order interactions.
In this experiment, the MBRL algorithm closely resembles the action selection of a repeated Lie bracket with some sampling noise -- recall that the controller only runs with a planning horizon of $\tau=5$.
We compare the actions selected by the MBRL algorithm and a Lie bracket sequence repeating each action 5 times ($\epsilon=0.05$) in \fig{fig:yaw-actions}, showing the ability for a data-driven approach to closely mimic a carefully derived analytical controller.

\section{Discussion \& Future Work}
\label{sec:results}
A model-based model-predictive controller learned to achieve a maximum yaw rate of up to \SI{31}{degrees / \second} by closely mimicking the Lie bracket control law.
It is important to note that the MBRL controller achieves this yaw rate in a full trial, and does not cause diverging attitude stability like that of the high-$\epsilon$ Lie bracket controllers.
With assembly variation and random initial states, the Lie bracket controller reaches the simulator stop condition (pitch or roll above \SI{45}{degrees} represents an unrecoverable state) in over $90\%$ of trials ($N=100$) before reaching an episode length of 1000 points (\SI{10}{\second}) -- a secondary controller preventing divergence is needed for practical application.
The MBRL MPC can maintain stability and yaw actuation in the presence of assembly and environmental uncertainty, and would not need other controllers to integrate future tasks. 

The means by which controllers are deployed is important to the practicality of novel methods, providing additional points of comparison for the use of the Lie-bracket or MBRL controllers with a real ionocraft.
The Lie-bracket controller would be one level in a hierarchical feedback controller, requiring more expert design, and the MBRL algorithm would need a modification of the cost function to include time-varying trajectory goals.
Translating the MBRL MPC to use onboard computation is an important avenue of future work, which is being explored by using advances in imitation and supervised learning to distill the MPC to a neural network policy -- which should be capable of running well above \SI{100}{\hertz} without a GPU.
The ionocraft's movement in the yaw axis is currently constrained by the power tethers required for flight, delaying application of our method to the experimental platform~\cite{drew2018toward}.
Other sources of variation, such as different forces per thruster, as modeled by $\beta_k$ in \eq{eq:thrust}, airframe flexibility, and robot-specific drag models, are not studied and could further incentivize data-driven techniques for this robot.
Integrating additional sources of uncertainty into the simulator would improve fidelity of both the Lie bracket and the MBRL.
\section{Conclusion}
\label{sec:concl}
This paper provided a case study in synthesising a yaw controller for a novel flying robot, the ionocraft.
Generating complex, nonlinear controllers on costly robots is a trade-off between engineer time and experimental cost.
Lie brackets, and other closed-form nonholonomic controllers, require non-trivial dynamics model understanding, which is a level of expertise often lacking when deploying robotic systems.
In this paper we characterized how a proven MBRL approach can closely match a derived controller at the low-level of action selection and in performance.
While translating the nonholonomic control law or model-based RL to real experiments poses high risk for different reasons -- namely a trade off between expert time and costly environmental interactions --  this paper shows that an encouraging new path exists in data-driven control for roboticists that is not dependant on familiarity with a new platform.

\section*{Acknowledgment}
The authors would like to thank Claire Tomlin, Francesco Borrelli, and Roberto Calandra for helpful discussions and the Berkeley Sensor \& Actuator Center for their support.
Thank you to Ryan Koh for helping with related code.
\bibliographystyle{IEEEtran}
\bibliography{main}  

\clearpage
\section*{Appendix}

\subsection{Lie Bracket Derivation}
Begin with any drift-free, control-affine nonlinear system with two inputs. 
\begin{equation}
    \dot{x} = f(x)u_1 + g(x) u_2
    \label{eq:example}
\end{equation}
With two vector fields $f, g$ execute the action sequence in \eq{eq:lie-control}.
The goal is to derive a third manifold of motion, as the time-step $\epsilon \rightarrow 0$
When applying a constant single-input to the system shown in \eq{eq:example}, the system takes the form of 
\begin{equation}
    \dot{x}=h(x)
    \label{eq:nonlinear}
\end{equation}

For use in the Taylor expansion, take the derivative with respect to time of the general form shown in \eq{eq:nonlinear}.
\begin{equation}
    \ddot{x}=\frac{dh(x)}{dt} = \frac{dh}{dx} \frac{dx}{dt} = \frac{dh}{dx}\dot{x} = \frac{dh}{dx}h(x)
    \label{eq:chain-rule}
\end{equation}
The derivative of the vector field $h(x)$ is Jacobian matrix, $\mathbb{J}\in \mathbb{R}^{n\times n}$ define by as
\begin{equation}
    \mathbb{J}\big(h(x)\big) = \frac{dh}{dx} = \begin{bmatrix}
    \frac{dh_1}{dx_1} & \frac{dh_1}{dx_2} & \cdots & \frac{dh_1}{dx_n} \\
    \frac{dh_2}{dx_1} & \frac{dh_2}{dx_2} & \cdots & \frac{dh_2}{dx_n} \\
    \vdots & \vdots & \ddots & \vdots \\
    \frac{dh_n}{dx_1} & \frac{dh_n}{dx_2} & \cdots & \frac{dh_n}{dx_n} \\
    \end{bmatrix}.
\end{equation}
The Lie-bracket emerges from an interweaving of two vector fields in a Taylor Expansion.
Consider the Taylor series in one dimension of the first vector field direction, $f$, with a time step of $\epsilon$.
\begin{align}
    x(\epsilon) & = x(0) + \epsilon \dot{x}(0)+\frac{1}{2}\epsilon^2 \ddot{x}(0) + \cdots \\
    & = x(0) + \epsilon f\big(x(0)\big)+\frac{1}{2}\epsilon^2 \frac{df}{dx}\Bigg|_{x(0)} f\big(x(0)\big)+ \cdots
    \label{eq:t1}
\end{align}
Now, consider application of the second vector field $g$ from $t\in[\epsilon,2\epsilon)$.
\begin{align}
    x(2\epsilon) & = x(\epsilon) + \epsilon \dot{x}(\epsilon)+\frac{1}{2}\epsilon^2 \ddot{x}(\epsilon) + \cdots \\
    & = x(\epsilon) + \epsilon g\big(x(\epsilon)\big)+\frac{1}{2}\epsilon^2 \frac{dg}{dx}\Bigg|_{x(\epsilon)} f\big(x(\epsilon)\big)+ \cdot
    \label{eq:t2}
\end{align}
Substituting \eq{eq:t1} into \eq{eq:t2} and dropping the $\big( x(0)\big)$ from every occurrence of $f,g$ yields an expression for $x(2\epsilon)$.
\begin{align}
    x(2\epsilon) & = x(0) + \epsilon (f+g)+\epsilon^2\Big(\frac{1}{2} \frac{df}{dx}f +\frac{dg}{dx}f+ \frac{1}{2} \frac{dg}{dx}g \Big)+ \cdots
\end{align}
Repeat the step of substituting the Taylor expansion of the previous vector field, for a time-step $\epsilon$ again.
\begin{align}
    x(3\epsilon) & = x(0) + \epsilon g+\epsilon^2\Big(\frac{dg}{dx}f-\frac{df}{dx}g +\frac{1}{2}\frac{dg}{dx}g \Big)+ \cdots
\end{align}
Finally, finish with the final substitution to obtain the estimate for $x(4\epsilon)$.
\begin{align}
    x(4\epsilon) & = x(0) +\epsilon^2\Big(\frac{dg}{dx}f - \frac{df}{dx}g \Big)+ \cdots 
\end{align}
Take the limit in a small time step to achieve the Lie-bracket vector field.
\begin{equation}
    \lim_{\epsilon\rightarrow 0} \frac{x(4\epsilon)-x(0)}{\epsilon^2} = \frac{dg}{dx}f - \frac{df}{dx}g 
\end{equation}
The Lie bracket vector field is defined as 
\begin{equation}
   \big[ f,g](x)  = \Big(\frac{dg}{dx}f - \frac{df}{dx}g\Big) (x)
\end{equation}
Which yields the state at time $t=4\epsilon$:
\begin{equation}
    x(4\epsilon) = x_0 + \epsilon^2 [f,g](x_0)+\mathcal{O}(\epsilon^3)
\end{equation}

\end{document}